\documentclass{svproc}
\usepackage{url}
\usepackage{comment}
\usepackage{siunitx}
\usepackage{amsmath}
\usepackage{multirow}
\usepackage{tikz} 
\usepackage{pgfplots}
\usepackage{hyperref}
\usepackage[export]{adjustbox}
\usepackage[capitalise]{cleveref}
\usepackage[xindy]{glossaries} 

\newacronym{res}{RES}{renewable energy resources}
\newacronym{pv}{PV}{photovoltaic}
\newacronym{ml}{ML}{machine learning}
\newacronym{nwp}{NWP}{numerical weather prediction}
\newacronym{ai}{AI}{artificial intelligence}
\newacronym{rgb}{RGB}{red–green–blue}
\newacronym{ghi}{GHI}{global horizontal irradiance}
\newacronym{dhi}{DHI}{diffuse horizontal irradiance}
\newacronym{dni}{DNI}{direct normal irradiance}
\newacronym{r/b}{R/B}{red–to–blue ratio}
\newacronym{r-b}{R-B}{red–to–blue difference}
\newacronym{r}{R}{red}
\newacronym{g}{G}{green}
\newacronym{b}{B}{blue}
\newacronym{knn}{kNN}{k-nearest neighbour}
\newacronym{saci}{SACI}{smart adaptive cloud identification} 
\newacronym{sic}{SIC}{smart image categorisation} 
\newacronym{ftm}{FTM}{fixed threshold method} 
\newacronym{mce}{MCE}{minimum cross entropy} 
\newacronym{csl}{CSL}{clear sky library} 
\newacronym{(r-b)/(r+b)}{(R-B)/(R+B)}{normalized RBR} 
\newacronym{rbr-csl}{RBR-CSL}{difference image} 
\newacronym{hyta}{HYTA}{hybrid thresholding technique} 
\newacronym{hcf}{HCF}{haze correction factor} 
\newacronym{rbr-(csl)(hcf)}{RBR-(CSL)(HCF)}{difference hcf} 
\newacronym{svc}{SVC}{support vector classification} 
\newacronym{asm}{ASM}{angular second-moment} 
\newacronym{csi}{CSI}{clear sky index k*} 
\newacronym{openCV}{OpenCV}{open source computer vision} 
\newacronym{cmvs}{CMVs}{cloud motion vectors} 
\newacronym{xgboost}{XGBoost}{random tree gradient boosting}
\newacronym{bma}{BMA}{block-matching algorithm}

\newacronym{nrel}{NREL}{national renewable energy laboratory}
\newacronym{rmse}{RMSE}{root mean square error}
\newacronym{nrmse}{NRMSE}{normalized root mean square error}
\newacronym{mae}{MAE}{mean absolute error}
\newacronym{mbe}{MBE}{mean bias error}
\newacronym{fs}{FS}{forecast skill}
\newacronym{pr}{PR}{precision}
\newacronym{re}{RE}{recall}
\newacronym{fov}{FOV}{field of view}
\newacronym{osmg}{OSMG}{oahu solar measurement grid}
\newacronym{surfrad}{SURFRAD}{surface radiation}
\newacronym{nsrdb}{NSRDB}{national solar radiation database}
\newacronym{soda}{SoDa}{solar radiation data}
\newacronym{srtm}{SRTM}{shuttle radar topography mission}
\newacronym{szas}{SZAs}{solar zenith angles}
\newacronym{tsi}{TSI}{total sky imager}
\newacronym{ir}{IR}{infrared radiation}
\usetikzlibrary{calc}
\usetikzlibrary{shapes.arrows}
\usetikzlibrary{positioning,fit,backgrounds}
\usetikzlibrary{shapes.geometric, arrows}
\usepackage{caption}  
\usepackage[normalem]{ulem}

\newlength{\posttableskip}
\usepackage{epstopdf}
\usepackage{floatrow}
\DeclareFloatFont{tiny}{\tiny}
\AppendGraphicsExtensions{.gif}

  \usepackage{eso-pic}
  \usepackage{color}
  \definecolor{DisclaimerGray}{gray}{0.92}

  \AddToShipoutPicture*{\put(23,755){\colorbox{DisclaimerGray}{\parbox{1.61\textwidth}
  {\footnotesize \copyright Springer, 2021. This is the author's version of the
  work. Personal use of this material is permitted. However, permission to
  reprint/republish this material for advertising or promotional purpose or for
  creating new collective works for resale or redistribution to servers or lists,
  or to reuse any copyrighted component of this work in other works must be
  obtained from the copyright holder. The definite version will be published within the FICC 2021 proceedings in the Springer series "Advances in Intelligent Systems and Computing", 2021.}}}}

\begin{document}
\mainmatter              
%

\title{A review on physical and data-driven based nowcasting methods using sky images}


\titlerunning{A review on nowcasting methods}  
%
\author{Ekanki Sharma and Wilfried Elmenreich}
%
%
%
\institute{{Institute of Networked and Embedded Systems \\ Lakeside Labs \\ Alpen-Adria-Universit\"{a}t\\
Klagenfurt, Austria\\}}

\maketitle              

\begin{abstract}

Amongst all the \gls{res}, solar is the most popular form of energy source and is of particular interest for its widely integration into the power grid. However, due to the intermittent nature of solar source, it is of the greatest significance to forecast solar irradiance to ensure uninterrupted and reliable power supply to serve the energy demand. There are several approaches to perform solar irradiance forecasting, for instance satellite-based methods, sky image-based methods, machine learning-based methods, and numerical weather prediction-based methods. In this paper, we present a review on short-term intra-hour solar prediction techniques known as nowcasting methods using sky images. Along with this, we also report and discuss which sky image features are significant for the nowcasting methods. 

\end{abstract}
\section{Introduction}\label{sec:introduction}

The transition from conventional energy source (e.g. fossil fuels) to sustainable energy source (e.g. renewable energy) is gaining momentum. This involves integrating \gls{res} in the power grid which contributes a viable solution to meet the increasing future energy demand. 
It thus helps in reducing the carbon footprint $(CO_{2})$ and greenhouse gases in the atmosphere. 
As long as photovoltaic systems are deployed in areas with sufficient solar yield, they have low impact on surroundings with respect to their energy production and they provide affordable energy since the fuel is free of cost \cite{Kahl2019}.
As a result of this transition, the PV installation is expected to increase by more than 4 \SI{}{\tera\watt} by 2025 and 21.9 \SI{}{\tera\watt} by 2050 \cite{Waldau2019}. 
However, integration of solar resource creates complex problems due to its intermittent nature in the energy management and scheduling. Renewable energy sources like photovoltaic systems typically rely on the weather and thus lead to variable energy production~\cite{sobe:informatik12}.
The integration also risks to the secure operation since it can cause an imbalance in demand and supply equilibrium. 
Estimating and forecasting solar irradiance has the potential to tackle these issues and reduce the integration and operation costs. 
The grid integration takes place at a variety of time scales which involves series of power system operations \& forecast submission requirements that varies accordingly. 
Therefore, forecast is performed over different time horizons corresponding to a particular decision making activity \cite{Wan2015}. 
The forecasting horizon can be defined as the period of time forecasts are made for. 
Forecasting horizons are categorised into very short-term, short-term, medium-term, and long-term forecasting. 
Very short-term forecasting is used for smart grid planning with a prediction period ranging from seconds to 30 \SI{}{\minute} \cite{REN2015}. 
This type of forecast is useful for utilities for determining electricity pricing and monitoring of real-time electricity dispatch \cite{AHMED2020}. Short-term forecasting is carried out for the prediction period ranging from 30 min to several hours \cite{REN2015}. 
This forecast is beneficial to the electricity market for making decisions like economic load dispatch and power system operation. 
Medium-term forecasting spans from 6 - 24h \cite{REN2015} and it is essential for maintenance scheduling of a power system integrated with energy sources. Long-term forecasting lasts till one week \cite{REN2015} which is suitable for long-term power generation, transmission, distribution \cite{BEHERA2018}. 
The increase in forecasting horizon degrades the forecasting accuracy \cite{GARCIAMARTOS2011}. 
A possible approach is to nest forecasting algorithms with different time horizons, thus using short-term forecasting methods for hourly forecasting and extending it then to medium-term and long-term forecasting by switching methods.
Filik et al.~in \cite{Filik2009} present a unified model for hourly load forecasting in short-, medium-, and long-terms with hourly accuracy. Other than these three forecasting horizons discussed above, there are further classification namely intra-hour, intra-day and day ahead. 
Intra-hour also known as nowcasting involves forecast horizons from seconds to an hour and this overlaps with very short-term and short-term which helps ensuring grid quality and stability. 
Island grids and low quality power supplies rely on such predictions~\cite{ZHANG2018}.

There are several approaches reported in the literature which aim at solar irradiance forecasting. 
They are broadly categorised into approaches based on satellite or sky images and data-driven methods including \gls{ai} and \gls{nwp}-based methods. 
This paper focuses on reviewing the techniques used for short-term intra-hour solar irradiance forecasting using sky images integrated with historical data. 
Along with this, the paper also reports the image features extracted from sky images which are significant for nowcasting.
To the best of our knowledge, there is no study present in the literature which reports a detailed review on the significant features or parameters extracted from sky images required for solar irradiance forecasting. 
This paper aims to close this research gap and to make the readers aware of the recent developments in this domain. 
The aim is also to make readers acquainted with the available datasets. \\  
The paper is organised as follows: \cref{sec:sky images based forecasting} addresses and discusses the related work on feature extraction and classification techniques applied on sky images for estimating solar irradiance. \Cref{sec:datasets}  reviews openly available datasets. \Cref{sec:Error metrics} discusses error metrics for the evaluation of the quality of an approach, followed by a conclusion of the paper and future work in \cref{sec:conclusion and future work}. 

\section{Sky images based forecasting}\label{sec:sky images based forecasting}

\Gls{pv} output power can be estimated either using indirect or direct approaches. 
The indirect approach assesses PV power output using the solar irradiance forecasts obtained using irradiance data along with other meteorological features. In contrast, direct methods rely on the output power of the PV for estimation. 
The work reported in this paper focuses on the indirect forecast approach where the aim is to address the techniques used for solar irradiance forecasting using sky-images with high spatial and temporal resolution. 
Spatial resolution refers to the size of one pixel which stands for a picture element i.e. the smallest individual 'block' that makes up the image. 
With higher resolution it is possible to distinguish smaller details in an image. 
Temporal resolution refers to the frequency with which the data is collected i.e. the resolution of a measurement with respect to time. 
Since satellite-based images suffer from poor spatial and temporal resolutions, images from sky cameras are recognised as a more capable solution for intra-hour solar irradiance forecast. 
 
Models based on sky images can be explored under two frameworks. 
The first framework is a physics-based model that typically follows a step by step approach as mentioned below. 

\begin{enumerate}
    \item discrimination between clear-sky pixels and cloudy pixels
    \item cloud classification and cloud optical depth determination
    \item determination of cloud motion and cloud advection
    \item cloud height and cloud shadow tracking
\end{enumerate}
     
\begin{figure}[ht]
    \centering
    \includegraphics[width=11cm]{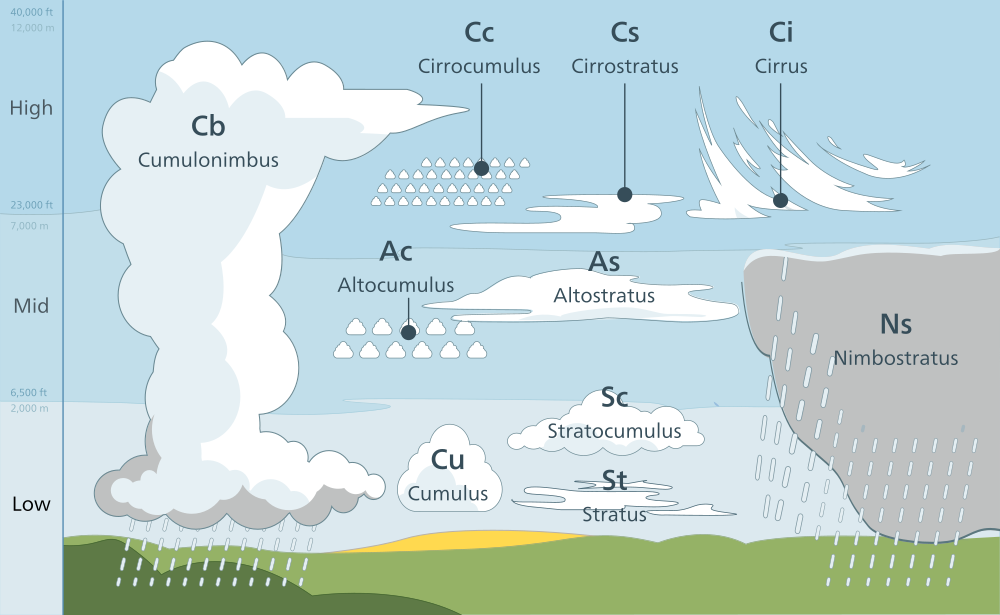}
    \caption{Cloud types (Image by Valentin de Bruyn / Coton under CC BY-SA 3.0)}
    \label{label:cloudtypes}
\end{figure}

The first step is to distinguish the pixels on the image between cloudy and clear-sky, followed by further classifying the images with cloudy pixels in different categories of cloud types with the corresponding layer according to altitude range (shown in \cref{label:cloudtypes}). 
The cloud classes along with their description according to International Cloud Classification System (ICCS) are shown in the table 1 below:

\begin{table}
\small
\begin{center}
\caption{Cloud types with description}
\begin{tabular}{lcp{.5em}p{5.8cm}}
\\
\hline\noalign{\smallskip}
Cloud levels & Cloud types & & Description \\
\noalign{\smallskip}\hline\noalign{\smallskip}
\multirow{3}{*}{High (5-13 km)} & Cirrus  & & Thin clouds, wispy \\
& Cirrocumulus &  & Fish scales like structure, whitish \\ 
& Cirrostratus &  & Sheet-like appearance, light grey to whitish \\ \hline
\multirow{3}{*}{Mid (2-7 km)} & Altocumulus  & & Globular patches of clouds, grayish-white color\\
& Altostratus &  & Uniformly grey, smooth \\ 
& Nimbostratus &  & Thick clouds, overcast, grey \\ \hline
\multirow{3}{*}{Low (0-2 km)} & Stratus  & & Thin layer clouds, grey, usually overcast\\
& Cumulus  & & Puffy clouds with well-defined rounded edges, white \\ 
& Cumulonimbus &  & Puffy thick clouds, mostly overcast, grey\\
& Stratocumulus & & Lumpy layer of clouds, broken to overcast, white or grey\\
\noalign{\smallskip}\hline
\end{tabular}
\end{center}\vspace{\posttableskip}
\end{table}

To assess cloud type classification, image sets are used which includes all the predefined cloud types. The selection procedure is performed sometimes manually by visually identifying the cloud type or sometimes according to phenomenological classes defined by international cloud classification system (as chosen by Heinle et al.~in \cite{Heinle2010}). 
Then, a cloud type classification algorithm for example \gls{knn} is applied. 
The last step calculating the cloud height and cloud shadow tracking is performed when multiple cameras are available. 

The second approach is based on a data-driven approach that relies on extraction of image features that are used as predictors in machine learning algorithms such as convolution neural network. These features include average, standard deviation, and entropy of images.  
Contributions by researchers in both approaches are covered in this paper.

A basic block diagram for sky images based solar irradiance forecast is shown in \cref{label:basicblocks}. 

\begin{center}
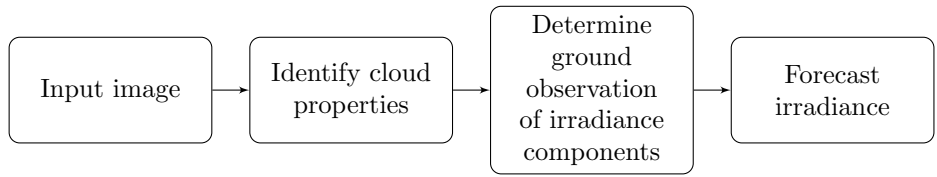

\centering
\tikzstyle{block} = [rectangle, draw, fill=white!50,   
    text width=7em, text centered, rounded corners,node distance=3.2cm, minimum height=4em]
\tikzstyle{line} = [draw, -latex']  
\begin{tikzpicture}
[node distance = .8cm, auto] 
    \node [block] (init) {Input image};  
    \node [block, right of= init](Identify cloud properties){Identify cloud properties}; 
    \node [block, right of= Identify cloud properties](Determine ground observation of irradiance components){Determine ground observation of irradiance components}; 
    \node [block, right of= Determine ground observation of irradiance components](Forecast irradiance){Forecast irradiance}; 
    \path [line] (init) -- (Identify cloud properties);   
    \path [line] (Identify cloud properties) -- (Determine ground observation of irradiance components);   
    \path [line] (Determine ground observation of irradiance components) -- (Forecast irradiance); 
\end{tikzpicture} 
\captionof{figure}{Basic building blocks}
\label{label:basicblocks}
\end{center}      

The first step is to acquire images using a sky camera. Typically, the sky images are acquired using an inexpensive upward looking camera equipped with fisheye lens to capture the whole sky image covering a \SI{180}{\degree} \gls{fov}.
Followed by this is a preprocessing step, where the cloud properties are identified which includes cloud coverage, cloud types using cloud classification techniques, cloud height, cloud velocity. The next step is to determine the ground measurements which includes \gls{ghi} and \gls{dhi}. \Gls{dni} is computed using a fundamental relationship between GHI, DHI and DNI given by this formula. Later irradiance forecast is performed using forecasting algorithm.

\begin{equation}
{\it GHI} = {\it DHI} + {\it DNI}\cdot \cos{\theta_{s}} 
\end{equation}

The original data in an image is provided in color. A partition into three components \gls{r}, \gls{g}, \gls{b} is made before features are calculated. 
Out of the three color components \gls{b} has the highest separation power due to the color of the sky and different translucency of clouds. The cloud-free sky shows higher value of \gls{b} intensity hence appear blue due to scattering of blue light by the air particles, whereas cloudy sky appear whitish or grey coloured due to the fact that \gls{b} and \gls{r} are scattered similarly.  

To estimate the cloud coverage, color features of an image are used. The color features describe the average color and tonal variation which help in distinguishing between thick dark clouds and bright clouds and also to separate high and transparent cirrus clouds from others.

The spectral features use the color components of the image to make a distinction between cloud types. However, considering only spectral features is not enough for an accurate classification, since they do not provide information about the spatial distribution of color in an image. 
The cumuliform clouds and stratocumulus cloud have similar mean color values, hence can not be separated using only spectral features. This problem is addressed by grey scale images which are obtained by transforming the \gls{rgb} cloud image into a single-channel feature image and each pixel is classified on the basis of threshold value of feature. The textural features exploits the grey scale image to describe the texture of the image. \\

For total cloud coverage estimation, various algorithms based on image processing have been proposed and tested on images obtained from sky cameras. The cloud coverage is detected by determining threshold which is obtained by ratio of \gls{r} to \gls{b} of a pixel in an image. 
In \cite{Long2005}, pixels with threshold value greater than 0.6 were classified as cloudy and pixels with lower value were classified as cloud-free. Long et al.~in \cite{Long2005} compared the results obtained by applying thresholding technique with visual observations. Results indicated an uncertainty of 0.2 while estimating fractional sky cover. Both the methods obtained similar result for overcast conditions. Long et al. summarised the problem created by using thresholding technique in region with circumsolar (situated near the Sun) pixels due to high aerosol conditions and misdetecting thin clouds. The Sun was obscured by a shadow band. A correction method is proposed which computes the possible error in total cloud coverage calculation. This correction value increases with increasing zenith angle and also introduce slight error greater than \SI{1}{\%}. \\

Instead of using ratio of \gls{r} and \gls{b}, Heinle et al.~in \cite{Heinle2010} used difference \gls{r-b}. Using difference threshold still results in minor errors however, results obtained outperformed threshold ratio method. The optimal value of \gls{r-b} utilized in the work is 30.

Kazantzidis et al.~in \cite{KAZANTZIDIS2012} present a multi-color criterion applied on sky images to determine total cloud coverage, considering visible percentage of Sun. Results are compared with the ground based weather observations. A set of images is used under different solar zenith angle were used. Results indicated that using \gls{r-b} and \gls{r/b} produce errors in identifying broken or overcast clouds under large zenith angles. Using multicolor criterion (which includes R, G and B intensities) for identifying overcast or broken clouds results in better performance as compared to using \gls{r-b}.   Several classification algorithms for classifying cloud types from sky images are reported in state-of-the-art. \\

In \cite{Singh2005}, five feature extraction methods are employed in a classifier based on kNN and neural network to identify different cloud types. Feature extraction methods used in the work are autocorrelation, co-occurrence matrices, edge frequency, Law’s features and primitive length. Singh and Glennen in \cite{Singh2005} concluded that no single feature extraction method is suitable for recognising all classes.\\

For identifying different cloud types, Calbo and Sabburg in \cite{Calbo2008} used digital images. They used three kinds of features spectral, features based on Fourier transform of an image and features that need distinction between cloudy and sky pixels. The automatic classification method implemented in the work show an agreement of \SI{62}{\%} (for eight sky conditions) and \SI{76}{\%} (for five sky conditions obtained after merging some of sky conditions). \\

Heinle et al.~in \cite{Heinle2010} propose an automatic cloud classification algorithm to classify sky images in real-time based on spectral and textural features presented in table 2. The implemented classifier is based on supervised \gls{knn} algorithm due to its high performance, simple implementation and low computational complexity. To avoid systematic misclassifications some of the cloud types were merged based on visual similarity. Out of all the ten cloud types presented in table 1, seven different types are selected after merging some of the cloud patterns depending on availability of data: high thin clouds (cirrus and cirrostratus), cumuliform clouds (cirrocumulus and altocumulus), stratocumulus clouds, low cumuliform clouds, thick clouds (cumulonimbus and nimbostratus), stratiform clouds. In addition clear sky type also included in the testing set which includes images with clear sky and images with cloudiness below \SI{10}{\%}.
 
\vspace{-.5cm}\begin{table}
\small
\begin{center}
\caption{Features utilized for cloud classification algorithm}
\begin{tabular}{lc}
\\
\hline\noalign{\smallskip}
Features & Name \\
\noalign{\smallskip}\hline\noalign{\smallskip}
\multirow{8}{*}{Spectral} & Mean R  \\
& Mean B  \\ 
& Standard deviation B \\ 
& Skewness B \\ 
& Difference R-G \\ 
& Difference R-B \\ 
& Difference G-B \\  \hline
\multirow{4}{*}{Textural} & Energy B  \\
& Entropy B  \\ 
& Contrast B \\
& Homogenity B \\
& Cloud Cover \\
\noalign{\smallskip}\hline
\end{tabular}
\end{center}\vspace{\posttableskip}
\end{table}

Heinle et al. chose 12 spectral and textural features which contains the color information of the image and grey levels of the images. They computed the features and stored them with assigned cloud type in the set of preclassified images. Then the test was conducted on the test sample including random images.  
To classify the images a supervised, non parametric kNN is implemented to classify the images. The results indicate that the classification accuracy achieved is in the range of 75\SI{}{\%} to 88\SI{}{\%}. Results also show that kNN algorithm could successfully classify the two classes of clouds clear sky and cirrus using only first and second order statistic features in addition to the actual cloudiness information. Nevertheless, several errors were reported in the paper.
One of the error is caused by misinterpretation of pixels near the sun which can be resolved by the determination of the position of solar disk and its removal.
This can be accomplished by using geometrical features.
Another error is related to confusion between cirrus and cumulus cloud types, when the cloudiness amounts less than 30\SI{}{\%}.
This error can be ruled out using a hierarchical classification process based on the cloudiness.
Similar error exists between cumulus and high cumulus due to similarity in their colour and smooth transition in definition.
Also confusions present between last three classes due to frequent changeover from one to another class.
Heinle et al. suggested to perform an initial partitioning of images into smaller subimages and classify them in case subimages include enough information to assign image parts to a cloud class. 
However, they concluded that an improved algorithm must be used by including features other than the spectral and textural features used in the work.\\

An improved automatic cloud classification algorithm using \gls{knn} is presented in \cite{KAZANTZIDIS2012}.  
In addition to statistical color and textural features, solar zenith angle, the cloud coverage, the visible fraction of solar disk and the existence of raindrops in sky images are also considered. 
To detect the existence of raindrops, solar zenith angle, cloud coverage and the visible fraction of sky disk new metrics are introduced. 
A set of images is used for testing and training the cloud classifier. The images are selected by visually inspecting them and assigned to different categories of cloud types. Using average values for each of the 12 cloud classifications does not provide the optimal result, since clouds of one class may vary significantly due to the metrics used for classification.
To overcome this problem, subclasses of every cloud class are created on the basis of extra features (solar zenith angle, cloud coverage, the visible fraction of solar disk) used in the work, which affect greatly the distribution of light in sky. 
The results obtained show the accuracy of classifier ranges between 78\SI{}{\%} and 95\SI{}{\%} in detecting seven cloud types correctly. The average performance of the classifier reported is 87.9\SI{}{\%}. The proposed multi color criterion results in an improvement in estimation of broken and overcast cloudiness and for large zenith angles. 
The kNN classifier could classify cumulus cloud with 91.9\SI{}{\%} accuracy. The remaining 8.1\SI{}{\%} of images which were actually of cumulus cloud type got misclassified and got assigned to two other cloud types, in particular cirrus-cirrostratus (5.1 \SI{}{\%}) and cirrocumulus-altocumulus (3\SI{}{\%})
However, misclassification does exist corresponding to the images with only few clouds close to the horizon far from the camera site. Cirrus-cirrostratus clouds are successfully detected with 94.6\SI{}{\%}. The classifier results show quite low performance 78\SI{}{\%} on cloud type cirrocumulus-altocumulus, the remaining images got misclassified and distributed to other classes, i.e., sky cumulus (8.5\SI{}{\%}), cirrus-cirrostratus (8.4\SI{}{\%}) and stratocumulus(5.1\SI{}{\%}). The clear skies were successfully detected by the classifier with a classification accuracy of 95\SI{}{\%}. The remaining images correspond to the days with considerable high values of aerosol optical depth (AOD), an optical parameter which is not considered in the classification algorithm. 

The study in \cite{Chu2014} proposes an automatic \gls{saci} system developed to combat the glare caused by using low cost alternative to sky imagery for cloud condition identification and solar irradiance forecast. 
\gls{saci} system uses \gls{sic} algorithm which combines sky images and solar irradiance measurements to classify sky condition in three categories: clear, overcast and partly cloudy. 
The clear sky period considered is defined as a period of time when clouds do not obscure sun and the total sky coverage is less than 5\SI{}{\%}. 
Overcast period is defined as a period of time when the sun is obscured by clouds and the total sky coverage is higher than 90\SI{}{\%}. 
The remaining data points are defined as partly cloudy. 
\Gls{saci} uses \gls{ftm} for overcast images, \gls{csl} and \gls{ftm} for clear images, \gls{csl} and \gls{mce} for partly cloudy images. 
\Gls{ftm} is based on the fact that the cloud pixels (in RGB image) have higher red (R) intensity values than sky pixels. A normalized \gls{(r-b)/(r+b)} is used in this work which is robust to noise. Using NRBR avoids extremely large RBRs when pixel have very low blue intensities hence improves visual contrast.
\Gls{mce} is an adaptive thresholding method based on Otsu algorithm \cite{Otsu1979}. In this method, it is important to maintain the threshold within an interval so as to ensure satisfactory performance. 
The interval limits are estimated from training set. 
Once the threshold is determined, pixels with higher \gls{r/b} values than the threshold are marked as cloudy. 
The presence of glare makes it difficult for \gls{ftm} and \gls{mce} to classify the pixels as cloud (opaque clouds) since it depends highly on solar geometry and can be tackled by using classification method based on \gls{csl} proposed by Shields et al.~in \cite{Shields1993}. The \Gls{csl} method uses a database of clear-sky images which removes the geometric variation of clear-sky RBRs that depend on sun-pixel angle and the solar zenith angle. In addition, a fixed threshold method to identify opaque clouds through a comparison with clear sky background RBR library are presented.
To address the variations in the haze amount, Shields et al. proposed an improved adaptive thresholding technique in \cite{shields2010}.
Based on the two findings reported in the previous work, \cite{Li2011} propose a \gls{hyta} for cloud detection. 
It was found in \cite{Cazorla2009}, \cite{Huo2010} that the presence of aerosols modifies the \gls{r/b} ratio and can impact the performance of cloud classification algorithms. 
Taking into account the significance of aerosols variations in the cloud detection and classification, Ghonima et al.~in \cite{Ghonima2012} propose a dynamic thresholding technique based on \gls{csl} method.  
To remove geometric variation of clear sky caused by sun-pixel angle and solar zenith angle, RBR of input image is subtracted by reference clear sky RBR that corresponds to same zenith angle, resulting in a difference image \gls{rbr-csl}. Further to account the haze effect caused by the presence of aerosol, a \gls{hcf} is utilized to correct the images which is defined as \gls{rbr-(csl)(hcf)} before applying cloud identification. 
The \gls{sic} algorithm proposed in \cite{Chu2014} receives images and the most recent \gls{ghi} data then decides which cloud identification method has to be applied. \gls{sic} integrates \gls{hyta} (classifies images based on their NRBR) with CSIT method (which uses GHI time series to detect the type of sky image). 
The architecture of proposed system is presented in Figure 3. 
\begin{figure}[ht]
    \centering
    \includegraphics[width=10cm]{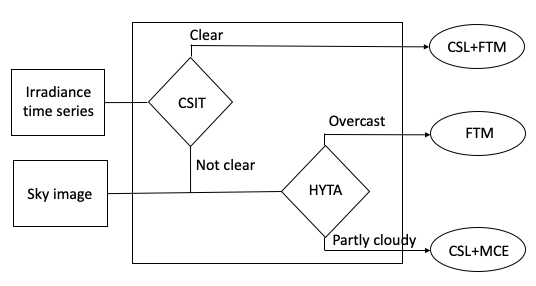}
    \caption{Architecture of proposed system}
\end{figure} 

A very short term \gls{ghi} prediction based on physics based models is presented in \cite{Schmidt2016}.
The data used in this study includes sky images, ceilometer-based cloud base height measurements and pyranometer data. 
The focus is to investigate the performance of the forecasting model under different cloud conditions. 
The cloud detection scheme considers binary states (sky/cloud).
In order to determine and predict the \gls{ghi} distribution from sky images, the preprocessing steps are employed on the images.
Schmidt et al. use the steps for sky image analysis and irradiance forecast shown in Figure 4. A modified \gls{r/b} for each pixel at the image position is proposed to overcome misclassification of circumsolar area. It also prevents the misclassification of thick and dark clouds caused by applying global threshold to the image. Static artificial objects are masked out from field of view in an image masking step. In cloud mapping step 3D position of cloudy pixel is determined using cloud base height obtained from ceilometer. 
\begin{center}
\centering
\begin{tikzpicture}[mynode/.style = {rectangle, draw, align=center,
            text width=6em,text centered,fill=white,
            inner xsep=4.5mm, inner ysep=3mm, rounded corners,node distance=6em},
            my arrow/.style={single arrow, draw,minimum height=.9cm},
rotate border/.style={shape border uses incircle, shape border rotate=#1},
            font=\sffamily]
    \node[mynode, label={[name=lab]Image analysis}] (inner1) {Input raw image};
    \node[mynode, below=2mm of inner1] (justbelow1) {Cloud detection};
    \node[mynode, below=2mm of justbelow1] (justbelow1a) {Image undistortion};
    \node[mynode, below=2mm of justbelow1a] (justbelow1b) {Cloud mapping};
    \node[mynode, below=2mm of justbelow1b] (justbelow1c) {Shadow mapping};
    
    \node[right=1.35cm of inner1, yshift = -28mm, mynode, label={[name=cal]Irradiance analysis}] (inner2) {Surface irradiance};
    \node[right=1.35cm of inner2,  yshift = 7mm, mynode, label={[name=img]Irradiance forecast}] (inner3) {Cloud motion};
    \node[mynode, below=2mm of inner3] (justbelow3) {Surface irradiance};
  
    \begin{scope}[on background layer]
    \node[fit={(lab) (inner1) (justbelow1c.south-|inner1.south)}, draw,fill=blue!20] (outer1) {};
    \node[fit={(cal) (inner2) (inner2)}, draw,fill=blue!20] (outer2) {};
    \node[fit={(img) (inner3) (justbelow3.south-|inner3.south)}, draw,fill=blue!20] (outer3) {};
    \end{scope}
    \path (outer1) -- (outer2) node[pos=0.45,yshift = -1mm,my arrow]{};
    \path (outer2) -- (outer3) node[pos=0.45,yshift = -2mm,my arrow]{};
\end{tikzpicture}

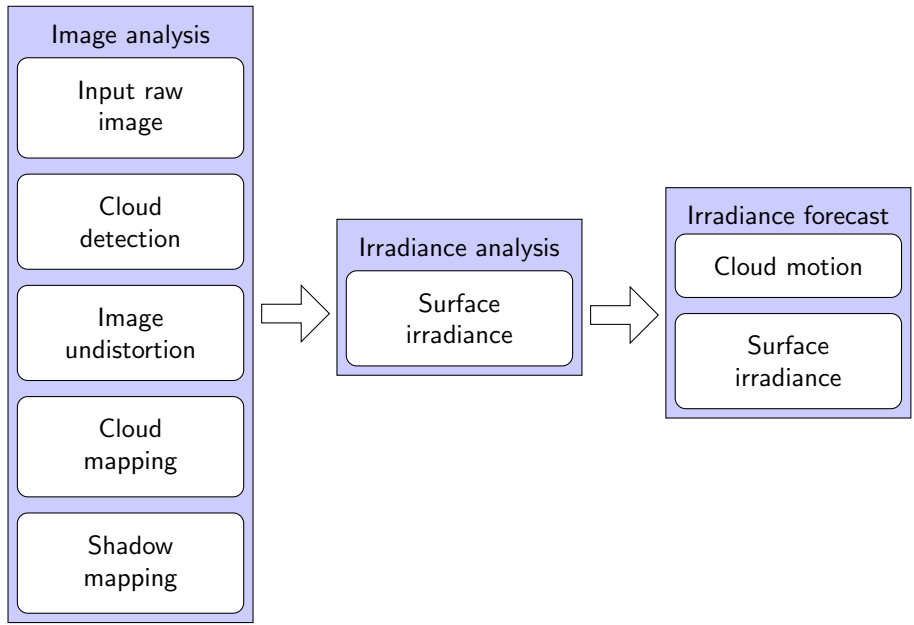
\captionof{figure}{Workflow for image analysis \& \gls{ghi} forecast}
\end{center}

The information obtained from the previous steps about the current sun position (azimuth angle and zenith angle) and cloud base height, the sun ray tracing is applied to map the cloud layer as a shadow layer on the ground. 
Cloud classification is applied further to classify each image instance in different cloud condition categories for evaluating forecast performance under different cloud conditions.  
Unlike the technique employed in previous work which is based on kNN, in this work classification technique based on \gls{svc} is applied. 
The number of features utilized for cloud classification algorithm include three extra features as compared to the ones used in previous work.  
All of these additional features include image texture properties derived from grey-level co-occurrence matrix defined by \cite{haralick1973}. The correlation is a measure of grey-tone linear dependencies, \gls{asm} is a measure of homogenity and dissimilarity is the measure of variation in grey level pairs in an image.

The next block performs irradiance analysis. The \Gls{csi} is a widely used term in solar irradiance forecasting which is defined as the ratio of measured irradiance to the calculated clear-sky irradiance. The formula is given below.  
The histogram corresponding to this measure indicates the presence of shadow (overcast condition) and no-shadow (clear sky condition) state. 

\begin{equation}
k* = \frac{GHI_\mathrm{meas}}{GHI_\mathrm{clear}}
\end{equation}
The corresponding GHI is computed as 
\begin{equation}
GHI = \frac{k*_\mathrm{hist}}{GHI_\mathrm{clear}}
\end{equation}

For performing the irradiance forecast, a fundamental information required is cloud movement and transformation. In the following we focus on cloud movement.
The cloud movement is obtained by applying optical flow algorithm available in \gls{openCV} to the original greyscale image, where the artificial objects are masked out. 
Applying optical flow algorithm results in \gls{cmvs}.  
The final step is to predict the solar irradiance upto 25 minutes ahead with an interval of every 15 seconds.
Two main questions are addressed in this work. 
First is concerning the accuracy of sky-imager-based analysis under different cloud conditions with respect to the distance from the camera and secondly its accuracy when compared with a persistence model. 
Irradiance analysis accuracy is evaluated on the basis of distance between pyranometer stations and camera and according to different cloud classes. The results indicate that the sky imager retrieval for distances of more than 1-2km from the camera under cumulus cloud conditions outperforms a single pyranometer measurement. An increase in distance is observed for stratocumulus and altocumulus/cirrocumulus to 2-3km and for nimbostratus/cumulonimbus to 6km. 
The overall forecast showed quite low performance compared to persistence. However, the increase in forecast performance is notable under heterogeneous cloud conditions, leading to increased variability in surface solar irradiance. \\

Two effective cloud discrimination methods from digital sky images using newly defined clear-sky index (CSI) are presented in \cite{Saito2016}. 
One method is an advanced method (AM) that uses RAW digital image format. 
The other method is a simplified method (SM) that uses digital signals in JPEG image format. 
The AM needs a RAW image, the spectral response functions in RGB channels, parameter \textit{s}, threshold of CSI for cloud discrimination, ozone column amount, and \gls{szas}. For determining parameter \textit{s} and CSI threshold, a set of training samples is created using different sky patterns. 
The sky patterns utilized for the work includes complete clear sky, optically thin cirrus clouds, thick broken clouds, and overcast clouds. 
The SM requires images in superfine mode in the JPEG format which has small file size (having three signals R,G,B in a pixel).
The results indicate that CSI is an important parameter of clouds. 
In addition to this, results also show that the correct classification rate from AM is better than that from SM, since using RAW images allow improved cloud discrimination. 
 
Marquez et al.~in \cite{Marquez2012} present relevance of using three sky cover indices derived from observed cloud cover via \gls{tsi}, \gls{ir} measurements and \gls{ghi} measurements for solar radiation modeling and forecasting. For solar radiation forecasting, a feed-forward ANN model composed of one input layer, one hidden layer and one output layer. To identify which time-series variables contribute to the best forecasting performance, four input combinations are applied as input to the ANN model. The first input combination used only SC index derived from \gls{ghi} measurements, the second set consists SC indices based on \gls{ghi} and \gls{ir}, the third set includes SC indices based on \gls{ghi} and \gls{tsi} followed by the fourth set consists of all the SC indices. The results indicate that using all the SC indices as input to the ANN model shows the highest forecasting accuracy with RMSE 35.9 \SI{}{\%}.  
  
In recent work \cite{Pedro2019A}, Pedro et al. present model which maps irradiance and image features directly into irradiance forecasts. Hence it avoids the steps involved in identifying cloud properties (cloud identification, cloud height, cloud velocity) which possibly can introduce error into the prediction outcome. 
A technique based on a \gls{bma}, to extract dynamic features from sky images to increase the accuracy of an intrahour forecasts for both \gls{ghi} and \gls{dni} values is employed. 
\Gls{bma} identifies the bulk motion of clouds relative to the position of the sun in the sky. 
An adaptive rectangular-shaped and wedge-shaped regions of interest are used for selecting the image pixels for the new features.
To produce an intrahour forecast for \gls{ghi} and \gls{dni}, \gls{xgboost} algorithms is used. 
The results are compared with a model using global features based on nonadaptive image features. 
An average increase of 6.8\SI{}{\%} and 6.7\SI{}{\%} in forecast skill for \gls{ghi} and \gls{dni} is achieved respectively when compared with model with nonadaptive image features. 
In comparison to clear-sky persistence, the model achieves forecast skills ranging from 20\SI{}{\%} to 30\SI{}{\%} for \gls{ghi} and 22\SI{}{\%} to 35\SI{}{\%} for \gls{dni}, which are the highest ever reported for these time horizons. 
An overall improvement in the performance metrics is obtained by applying feature engineering to extract information from the sky images. Feature extraction techniques applied are based on mutual information and Pearson correlation coefficients between image features and training data.
Most important is the simplicity of the model proposed in this work. The model directly maps irradiance and image features directly into the target forecasts avoiding the steps discussed in \cite{Pedro2019A} which includes cloud identifications, estimations of cloud velocity and height, since it introduce error in the final prediction of solar irradiance.
However, Pedro et al. conclude that \gls{bma} filtering method has to be extended by making modifications in the feature extraction algorithm to account for the cloud velocity. Including cloud motion information specifically in a region near the apparent position of the Sun, in adaptive sky image features do benefit solar irradiance forecast. An example for slow moving clouds, pixels near the sun have a larger weight than the pixel toward the image's edge and vice versa for fast moving clouds.

For nowcasting application, identifying and forecasting the solar position in the images, especially when covered by clouds is a key issue. Nespoli and Niccolai address this issue in \cite{Nespoli2020} and present three different techniques for identification of solar position in images. 
The first method is based on solar angles based identification which relies on trigonometric relations in order to find the sun position on the images. 
It works with determination of theoretical coordinates in terms of azimuth and solar elevation angle as a function of time. 
Once theoretical angles are computed, they are mapped to acquired images through trigonometric relations. 
The second method is based on image processing algorithms and is capable of identifying the sun by means of its shape and colour. 
The image is firstly masked for eliminating all the reflected areas that are not representative of sky conditions. This procedure can be easily performed because system shape does not change over time. Once the background is excluded, each color channel is analysed by means of a threshold value filtering technique, to locate the darker spot which is the sun. The method reported shows that it is very accurate and it require less computation time to process the images. However, it is difficult to locate the sun when it is covered by clouds. 
The last method is based on neural networks, which aims at solving drawbacks of both  previously mentioned techniques. ANN takes two inputs, the theoretical angle of the sun ($\alpha$ ,$\gamma$), and generates two outputs the $x-$ and the $y-$ coordinates of the sun on the image. The main drawback related to this methodology is the requirement of training set. It is important to cover a wide range of possible solar angles since neural networks are effective in interpolation. 

The three methodologies are compared on the same day. The ability of ANN-based method is highlighted since it is not affected by cloud coverage and it can estimate the sun position in early morning and late afternoon.  
Table 3 presents the forecasting models used for implementing nowcasting methods.  

\begin{table}
\small
\begin{center}
\caption{Nowcasting methods}
\begin{tabular}{lccp{1.0em}p{5.8cm}}
\\
\hline\noalign{\smallskip}
Method & Prediction horizon & Year \& Reference \\
\noalign{\smallskip}\hline\noalign{\smallskip}
ANN feed-forward model & 1 h  & 2012 \cite{Marquez2012} \\
Persistence model & 3-15 min & 2013 \cite{Marquez2013} \\ 
Deterministic \& MLP model & 5-15 min & 2016 \cite{Li2016} \\ 
Ineichen model \cite{Ineichen2008} & 5-15 min  & 2018 \cite{Bone2018}\\
\Gls{xgboost}   & 5-10 min & 2019 \cite{Pedro2019A} \\ 

\noalign{\smallskip}\hline
\end{tabular}
\end{center}\vspace{\posttableskip}
\end{table}

\section{Datasets}\label{sec:datasets}
Solar forecasting is an important technique that enables the integration of variable solar power generation into an electric grid. Despite of the growing interest in this domain, lack of standardised datasets is a big limitation in reproducing and benchmarking the forecasting models.
Secondly it is a limitation in progress particularly for those who do not have resources to obtain their own dataset. 
Recently some efforts have been made to close this research gap by facilitating access to the public data. 

An open-source tool is provided in \cite{Yang2018} for easy access to a publicly available dataset. 
The author provides an overview of five datasets along with code segments which can help in understanding how the datasets can be used in solar research. This further facilitates future contribution and collaboration. The datasets used for developing the tool are mentioned below:
\begin{enumerate}
    \item NREL \Gls{nsrdb}\footnote{\url{https://maps.nrel.gov/nsrdb-viewer/}} is a collection of hourly and half hourly values of various satellite derived irradiance dataset along with meteorological data. 
    \item NREL \Gls{osmg}\footnote{\url{https://midcdmz.nrel.gov/apps/sitehome.pl?site=OAHUGRID}} dataset includes global horizontal and tilted irradiance recorded every second. The data is available from 2010-03 to 2011-10.
    \item NOAA \Gls{surfrad}\footnote{\url{https://www.esrl.noaa.gov/gmd/grad/surfrad/dataplot.html}} records three components of solar irradiance along with several meteorological parameters such as ambient temperature, station pressure, wind speed. The SURFRAD network was established in 1993, commenced the operation with four stations in 1995. Later two more stations were added in 1998 followed by the seventh in 2003. Due to different commencing date, the length of data for each station varies. 
    \item The \gls{soda}\footnote{\url{http://www.soda-pro.com/web-services}} offers collection of paid and free solar radiation and solar related data.
    \item NASA \gls{srtm}\footnote{\url{http://https://dds.cr.usgs.gov/srtm/}} dataset provides worldwide altitude measurements which can be used for developing clear sky models.
\end{enumerate}

As discussed earlier the dataset \gls{surfrad} reported in \cite{Augustine2000} provides access to solar irradiance data and other meteorological features including wind speed and direction, air temperature, relative humidity and station pressure. The samples are recorded with resolution of 1 second.

The dataset reported in \cite{SENGUPTA2018}, \gls{nsrdb} consists of solar radiation derived from satellite with half-hourly resolution and meteorological data over the United States and the surrounding regions.
The work provides a comprehensive dataset which can be utilised for solar irradiance forecasting. 

Another addition to the availability of quality dataset as an open source includes a recent work reported in \cite{Pedro2019} which present a comprehensive dataset\footnote{\url{https://zenodo.org/record/2826939##.X5_5uUJKhTY}} that includes three years of quality controlled dataset for \gls{ghi} and \gls{dni} with one minute resolution. To develop and benchmark solar forecasting methods for intra-hour, intra-day and day ahead the dataset provides multiple years of detailed data of irradiance and weather, along with high resolution sky images, satellite images and \gls{nwp} data for the same target area and time interval. 
Authors also provide sample codes using simple regression methods to provide baseline models for future studies.
The sky camera captures Red-Green-Blue (RGB) color images at a medium resolution of 1536 $\times$ 1536 pixels with 1 minute interval.
The images can be used for both the frameworks namely physics based models and data-driven models. Open source datasets that report ground based measurements and sky images are summarised in Table 4.    

\begin{table}
\small
\begin{center}
\caption{Available datasets on sky images and ground measurements}
\begin{tabular}{p{2.4cm}p{3.6cm}p{.5em}p{3.7cm}r}
\\
\hline\noalign{\smallskip}
Dataset &  Data description & & Data resolution & Year \\
\noalign{\smallskip}\hline\noalign{\smallskip}
\Gls{osmg} &  Ground based irradiance & & 17 stations\newline 1 sec GHI Grid\newline 3 sec RSR 3-Component & 2010-2011 \\
\hline
\Gls{surfrad} & Ground based irradiance, meteorological data & & 7 stations \newline 3 min till 1 Jan 2009\newline 1 min since 1 Jan 2009& 1995-present\\
\hline
Comprehensive dataset & Ground based irradiance, sky images, NWP forecasts, satellite images & & 1 min & 2014-2016\\
\noalign{\smallskip}\hline
\end{tabular}
\end{center}\vspace{\posttableskip}
\end{table}

\section{Error metrics}\label{sec:Error metrics}
\subsection{Error metrics to evaluate solar irradiance forecasts}
For evaluating the accuracy of solar irradiance forecasting set of metrics are used. Out of all statistical metrics are widely used to conduct the performance evaluation of a forecasting model which includes \gls{mae}, \gls{mbe}, \gls{rmse} and \gls{nrmse}. Another important metric is \gls{fs} which is used to define the accuracy and degree of association of predicted observation over simplified historical observations. The definition of these metrics is given in the following. In addition, a detailed description can be found in \cite{Coimbra2013}, \cite{Zhang2015}. 

\begin{equation}
{\it MAE} = \sum_{i=1}^{N}\frac{|p_\mathrm{pred}-p_\mathrm{meas}|}{N}
\end{equation}

\begin{equation}
{\it MBE} = \sum_{i=1}^{N}\frac{ (p_\mathrm{pred}-p_\mathrm{meas})}{N}
\end{equation}

\begin{equation}
{\it RMSE} =  \sqrt {\frac{1}{N}\sum_{i=1}^N\left(p_\mathrm{pred}-p_\mathrm{meas}\right)^2}
\end{equation}

\begin{equation}
{\it FS} = 1- \frac{RMSE_\mathrm{pred}}{RMSE_\mathrm{meas}}
\end{equation}

\subsection{Error metrics to evaluate sky imagery based forecasts}
Typically evaluation of accuracy of sky image based forecast is conducted after converting the images into time series forecasts. However, for evaluating the accuracy of a classification algorithm used in classifying different cloud types while processing a sky image, an accuracy measure named ACC is utilized \cite{Schmidt2016}. 
ACC is defined as the ratio of correctly predicted states (sunny or cloudy) over all instances.

\begin{equation}
{\it ACC} = \frac{TS+TC}{TS+TC+FS+FC}
\end{equation}

To predict the visibility of the state of the sun authors in \cite{BERNECKER2014} used three metrics namely \gls{pr}, \gls{re}, $F_{\beta}$. 
\Gls{pr} refers to how well predicted occlusions match actual occlusion. \Gls{re} refers to how well the actual occlusions are predicted. $F_{\beta}$ score combines both of these values in a single metric. 

\begin{equation}
{\it PR} = \frac{hits}{hits+false alarms}
\end{equation}

\begin{equation}
{\it RE} = \frac{hits}{hits+misses}
\end{equation}

\begin{equation}
 F_{\beta} = (1+\beta^2)\frac{PR\cdot RE}{\beta^2\cdot PR+RE}
\end{equation}

\section{Conclusion and future work}\label{sec:conclusion and future work}

A detailed review on short-term intra-hour (nowcasting) solar irradiance forecasting using sky images has been carried out in this work. 
We reviewed and discussed the benefit of using ground based sky images with high spatial and temporal resolution in addition to weather data to estimate the solar irradiance.

The models based on physics and data-driven exist in the literature are discussed in this work, in order to make the readers aware of the recent developments in this domain. Furthermore, a brief discussion on available standardised datasets is also presented which can facilitate reproducibility and benchmarking of forecasting models. Followed by error metrics used for evaluating the performance of forecasting models and classification methods is also presented.  

We can draw several conclusions from the reported work. First we can infer that using low cost sky camera instead of sky imagers introduces excessive glare in the images, which in turn degrades the performance of classification algorithms. The image analysis in the preprocessing step is highly important to achieve better irradiance forecasting results. Using adaptive sky image features that depend upon the cloud motion in a region near the position of sun in the sky is beneficial for solar irradiance forecasting.

The classification algorithm based on \gls{svc} also showed promising results in achieving better classification of different cloud types. An increase in forecasting accuracy can also be achieved by replacing \gls{knn} by \gls{xgboost}. If more features are added to the set of predictors, it can boost the forecasting skill. Along with this identifying the solar position under clear and overcast conditions in sky images also proves to be beneficial. The method based on ANN turned out to be more accurate and reliable than the solar angle and image processing based identification methods. However, in order to generalize a trend it is important to cover entire range of possible solar angles. 

An important aspect for the research of forecasting methods is the availability of suitable and available datasets. In this review, we especially identified the dataset by Pedro et al.~\cite{Pedro2019} to be of interest. It is a comprehensive dataset supporting the development of hybrid models, since the dataset provides an access to multiple exogenous inputs including sky or satellite imagery, \gls{ghi} and \gls{dni} for the duration of three complete years.

Future work in nowcasting methods will require two types of contribution. First a proper comparison of prediction performances for a reference set of sky images is required. This should be done with open data and open source implementations of algorithms, in order to ensure reproducibility of results (cf.~\cite{elmenreich:repoducibility}). Second, based on existing algorithms, improved methods, possibly via hybrid models are expected to be created, once the problem of reproducible assessment is solved.


\begin{thebibliography}{6}
%

\bibitem{Kahl2019}
A.~Kahl, J{\'e}. Dujardin, and M.~Lehning.
\newblock The bright side of pv production in snow-covered mountains.
\newblock {\em Proceedings of the National Academy of Sciences},
  116(4):1162--1167, 2019.

\bibitem{Waldau2019}
A.~J. Waldau.
\newblock Snapshot of photovoltaics.
\newblock {\em Energies}, 12(5):7, Feb 2019.

\bibitem{sobe:informatik12}
A.~Sobe and W.~Elmenreich.
\newblock Smart Microgrids: Overview and Outlook.
\newblock In {\em Proceedings of the ITG INFORMATIK Workshop on Smart Grids}, Braunschweig, Germany, Sep 2012.

\bibitem{Wan2015}
C.~Wan, J.~Zhao, Y.~Song, Z.~Xu, J.~Lin, and Z.~Hu.
\newblock Photovoltaic and solar power forecasting for smart grid energy
  management.
\newblock {\em CSEE Journal of Power and Energy Systems}, 1(4):38--46, Dec
  2015.

\bibitem{REN2015}
Ye~Ren, P.N. Suganthan, and N.~Srikanth.
\newblock Ensemble methods for wind and solar power forecasting—a
  state-of-the-art review.
\newblock {\em Renewable and Sustainable Energy Reviews}, 50:82 -- 91, 2015.

\bibitem{AHMED2020}
R.~Ahmed, V.~Sreeram, Y.~Mishra, and M.D. Arif.
\newblock A review and evaluation of the state-of-the-art in pv solar power
  forecasting: Techniques and optimization.
\newblock {\em Renewable and Sustainable Energy Reviews}, 124:109792, 2020.

\bibitem{BEHERA2018}
M.~K. Behera, I.~Majumder, and N.~Nayak.
\newblock Solar photovoltaic power forecasting using optimized modified extreme
  learning machine technique.
\newblock {\em Engineering Science and Technology, an International Journal},
  21(3):428 -- 438, 2018.
  
\bibitem{GARCIAMARTOS2011}
C.~García-Martos, J.~Rodríguez, and M.~J. Sánchez.
\newblock Forecasting electricity prices and their volatilities using
  unobserved components.
\newblock {\em Energy Economics}, 33(6):1227 -- 1239, 2011.

\bibitem{Filik2009}
U.~{Basaran Filik}, O.~N. {Gerek}, and M.~{Kurban}.
\newblock Hourly forecasting of long term electric energy demand using a novel
  modeling approach.
\newblock In {\em 2009 Fourth International Conference on Innovative Computing,
  Information and Control (ICICIC)}, pages 115--118, 2009.

\bibitem{ZHANG2018}
J.~Zhang, R.~Verschae, S.~Nobuhara, and Jean-François Lalonde.
\newblock Deep photovoltaic nowcasting.
\newblock {\em Solar Energy}, 176:267 -- 276, 2018.
  
\bibitem{Heinle2010}
A.~Heinle, A.~Macke, and A.~Srivastav.
\newblock Automatic cloud classification of whole sky images.
\newblock {\em Atmospheric Measurement Techniques}, 3(3):557--567, 2010.

\bibitem{Long2005}
C.~N. Long, J.~M. Sabburg, J.~Calbó, and D.~Pagès.
\newblock {Retrieving Cloud Characteristics from Ground-Based Daytime Color
  All-Sky Images}.
\newblock {\em Journal of Atmospheric and Oceanic Technology}, 23(5):633--652,
  05 2006.

\bibitem{KAZANTZIDIS2012}
A.~Kazantzidis, P.~Tzoumanikas, A.F. Bais, S.~Fotopoulos, and G.~Economou.
\newblock Cloud detection and classification with the use of whole-sky
  ground-based images.
\newblock {\em Atmospheric Research}, 113:80 -- 88, 2012.
  
\bibitem{Singh2005}
M.~Singh and M.~Glennen.
\newblock Automated ground-based cloud recognition.
\newblock {\em Pattern Analysis and Applications}, 8(3):258–271, 2005.

\bibitem{Calbo2008}
J.~Calbo and J.~Sabburg.
\newblock Feature Extraction from Whole-Sky Ground-Based Images for Cloud-Type Recognition.
\newblock {\em Journal of Atmospheric and Oceanic Technology}, 25(1):3–14, 2008.

\bibitem{Chu2014}
Y.~Chu, H.~T.~C. Pedro, L.~Nonnenmacher, R.~H. Inman, Z.~Liao, , and C.~F.~M.
  Coimbra.
\newblock A smart image-based cloud detection system for intrahour solar
  irradiance forecasts.
\newblock {\em Atmospheric and Oceanic Techniques}, 31(9):1995–2007, 2014.

\bibitem{Otsu1979}
N.~Otsu.
\newblock A threshold selection method from gray-level histograms.
\newblock {\em IEEE Transactions on Systems, Man, and Cybernetics}, 9:62--66,
  1979.

\bibitem{Shields1993}
R.~W.~Johnson Shields, J.~E. and T.~L. Koehler.
\newblock Automated whole sky imaging systems for cloud field assessment.
\newblock {\em American Meteorological Society}, page 17–22, 1993.

\bibitem{shields2010}
J.E. Shields, M.E. Karr, A.R. Burden, R.W. Johnson, and W.S. Hodgkiss.
\newblock {\em Scientific Report on Whole Sky Imager Characterization of Sky
  Obscuration by Clouds for the Starfire Optical Range: Scientific Report for
  AFRL Contract FA9451-008-C-0226}.
\newblock University of California San Diego, Scripps Instiution of
  Oceanography, Marine Physical Lab.

\bibitem{Li2011}
Qingyong Li, Weitao Lu, and Jun Yang.
\newblock {A Hybrid Thresholding Algorithm for Cloud Detection on Ground-Based
  Color Images}.
\newblock {\em Journal of Atmospheric and Oceanic Technology},
  28(10):1286--1296, 10 2011.

\bibitem{Cazorla2009}
A.~Cazorla, J.~E. Shields, M.~E. Karr, F.~J. Olmo, A.~Burden, and
  L.~Alados-Arboledas.
\newblock Technical note: Determination of aerosol optical properties by a
  calibrated sky imager.
\newblock {\em Atmospheric Chemistry and Physics}, 9(17):6417--6427, 2009.

\bibitem{Huo2010}
J.~Huo and Daren L{\"u}.
\newblock Preliminary retrieval of aerosol optical depth from all-sky images.
\newblock {\em Advances in Atmospheric Sciences}, 27:421--426, 2010.

\bibitem{Ghonima2012}
M.~S. Ghonima, B.~Urquhart, C.~W. Chow, J.~E. Shields, A.~Cazorla, and J.~Kleissl.
\newblock A method for cloud detection and opacity classification based on
  ground based sky imagery.
\newblock {\em Atmospheric Measurement Techniques}, 5(11):2881--2892, 2012.

\bibitem{Schmidt2016}
T.~Schmidt, J.~Kalisch, E.~Lorenz, and D.~Heinemann.
\newblock Evaluating the spatio-temporal performance of
  sky-imager-based solar irradiance analysis and forecasts.
\newblock {\em Atmospheric Chemistry and Physics}, 16(5):3399--3412, 2016.

\bibitem{haralick1973}
{R.} Haralick, {K.} Shanmugam, and {I.} Dinstein.
\newblock Texture features for image classification.
\newblock {\em IEEE Transactions on Systems, Man, and Cybernetics}, 3(6), 1973.

\bibitem{Saito2016}
M.~Saito and H.~Iwabuchi.
\newblock Cloud discrimination from sky images using a clear-sky index.
\newblock {\em American Meteorological Society}, 33(8):1583–1595, 2016.

\bibitem{Marquez2012}
R.~Marquez, V.~Gueorguiev and C.~Coimbra.
\newblock Forecasting of global horizontal irradiance using sky cover indices.
\newblock {\em Solar Energy Engineering}, 135(1), 2012.

\bibitem{Marquez2013}
R.~Marquez and C.~Coimbra.
\newblock Intra-hour DNI forecasting based on cloud tracking image analysis.
\newblock {\em Solar Energy}, 91, pages 327-336, 2013.

\bibitem{Li2016}
M.~Li, Y.~Chu, H.~Pedro and C.~Coimbra.
\newblock Quantitative evaluation of the impact of cloud transmittance and cloud velocity on the accuracy of short-term DNI forecasts.
\newblock {\em Renewable Energy}, 86, pages 1362-1371, 2016. 

\bibitem{Ineichen2008}
P.~Ineichen.
\newblock A broadband simplified version of the Solis clear sky model.
\newblock {\em Solar Energy}, 82(8), pages 758-762, 2008. 

\bibitem{Bone2018}
V.~Bone, J.~Pidgeon, M.~Kearney and A.~Veeraragavan.
\newblock Intra-hour direct normal irradiance forecasting through adaptive clear-sky modelling and cloud tracking.
\newblock {\em Solar Energy}, 159, pages 852-867, 2018. 

\bibitem{Pedro2019A}
Hugo T.~C. Pedro, Carlos F.~M. Coimbra, and Philippe Lauret.
\newblock Adaptive image features for intra-hour solar forecasts.
\newblock {\em Journal of Renewable and Sustainable Energy}, 11(3):036101,
  2019.
  
\bibitem{Pedro2019}
Hugo T.~C. Pedro, David~P. Larson, and Carlos F.~M. Coimbra.
\newblock A comprehensive dataset for the accelerated development and
  benchmarking of solar forecasting methods.
\newblock {\em Journal of Renewable and Sustainable Energy}, 11(3):036102,
  2019.

\bibitem{Nespoli2020}
A.~{Nespoli} and A.~{Niccolai}.
\newblock Solar position identification on sky images for photovoltaic nowcasting applications.
\newblock In {\em 2020 IEEE International Conference on Environment and
  Electrical Engineering and 2020 IEEE Industrial and Commercial Power Systems
  Europe (EEEIC / I CPS Europe)}, pages 1--5, 2020.
  
\bibitem{Yang2018}
D.~Yang.
\newblock Solardata: An R package for easy access of publicly available solar
  datasets.
\newblock {\em Solar Energy}, 171, 2018.

\bibitem{Augustine2000}
Augustine, John A. and DeLuisi, John J. and Long, Charles N.
A.~Augustine, J.~DeLuisi, N.~Long.
\newblock SURFRAD–A National Surface Radiation Budget Network for Atmospheric Research.
\newblock {\em Bulletin of the American Meteorological Society}, 81(10),2341 -- 2358, 2000.

\bibitem{SENGUPTA2018}
M.~Sengupta, Y.~Xie, A.~Lopez, A.~Habte, G.~Maclaurin, and J.~Shelby.
\newblock The national solar radiation data base (NSRDB).
\newblock {\em Renewable and Sustainable Energy Reviews}, 89:51 -- 60, 2018.

\bibitem{Coimbra2013}
C.~F.~M. Coimbra, J.~Kleissl, and R.~C. M{\'a}rquez.
\newblock Chapter 8 – overview of solar-forecasting methods and a metric for
  accuracy evaluation.
\newblock 2013.

\bibitem{Zhang2015}
J.~Zhang, A.~Florita, B.~Hodge, S.~Lu, H.~F. Hamann, V.~Banunarayanan, and
  A.~M. Brockway.
\newblock A suite of metrics for assessing the performance of solar power
  forecasting.
\newblock {\em Solar Energy}, 111:157 -- 175, 2015.

\bibitem{BERNECKER2014}
D.~Bernecker, C.~Riess, E.~Angelopoulou, and J.~Hornegger.
\newblock Continuous short-term irradiance forecasts using sky images.
\newblock {\em Solar Energy}, 110:303 -- 315, 2014.

\bibitem{elmenreich:repoducibility}
W.~Elmenreich, P.~Moll, S.~Theuermann, and M.~Lux.
\newblock Making simulation results reproducible - {S}urvey, guidelines, and examples based on {G}radle and {D}ocker.
\newblock {\em PeerJ Computer Science}, 5(e240):1 -- 27, 2019.



\end{thebibliography}

\end{document}